# IN GENERATIVE AI WE TRUST: CAN CHATBOTS EFFECTIVELY VERIFY POLITICAL INFORMATION?

Elizaveta Kuznetsova[1], Mykola Makhortykh[2], Victoria Vziatysheva[3], Martha Stolze[4], Ani Baghumyan[5], Aleksandra Urman[6]

**ABSTRACT**. This article presents a comparative analysis of the ability of two large language model (LLM)-based chatbots—ChatGPT and Bing Chat (recently rebranded to Microsoft Copilot)—to detect veracity of political information. We use AI auditing methodology to investigate how chatbots evaluate true, false, and borderline statements on five topics: COVID-19, Russian aggression against Ukraine, the Holocaust, climate change, and LGBTQ+-related debates. We compare how the chatbots perform in high- and low-resource languages by using prompts in English, Russian, and Ukrainian. Furthermore, we explore chatbots' ability to evaluate statements according to political communication concepts of disinformation, misinformation, and conspiracy theory, using definition-oriented prompts. We also systematically test how such evaluations are influenced by source bias which we model by attributing specific claims to various political and social actors. The results show high performance of ChatGPT for the baseline veracity evaluation task, with 72% of the cases evaluated correctly on average across languages without pre-training. Bing Chat performed worse with a 67% accuracy. We observe significant disparities in how chatbots evaluate prompts in high- and low-resource languages and how they adapt their evaluations to political communication concepts with ChatGPT providing more nuanced outputs than Bing Chat. Finally, we find that for some veracity detection-related tasks, the performance of chatbots varied depending on the topic of the statement or the source to which it is attributed. These findings highlight the potential of LLM-based chatbots in tackling different forms of false information in online environments, but also points to the substantial variation in terms of how such potential is realized due to specific factors (e.g. language of the prompt or the topic).

## 1. INTRODUCTION

Artificial intelligence (AI)-driven systems have for long been recognized as crucial factors in shaping online political information environments (Blumler, 2016). Among other things, these systems are applied for automated information curation, a process of selecting and presenting content from a pull of data (Rader & Gray, 2015). Ranging from search engines to recommender systems, curation mechanisms pose multiple challenges for society

[1] Weizenbaum Institute for the Networked Society, Berlin, elizaveta.kuznetsova@weizenbaum-institut.de
[2] University of Bern, Bern
[3] University of Bern, Bern
[4] Weizenbaum Institute for the Networked Society, Berlin
[5] University of Bern, Bern
[6] University of Zurich, Zurich

(Fouquaert & Mechant, 2022; Rader & Gray, 2015): From affecting information flows (Unkel & Haas, 2017) to determining individual exposure to propaganda content (Kuznetsova & Makhortykh, 2023), recent studies have highlighted problems of algorithmic bias (Bonart et al., 2020; Bozdag, 2013), political microtargeting (Gräfe, 2019), content personalization mechanisms on digital platforms (Bastian et al., 2019), and disruptive content presence and mitigation (Aguerri & Santisteban, 2022).

The development of Large Language Models (LLMs), a form of AI technology capable of processing and generating textual content (Naveed et al., 2023), signifies a new stage in the complex relationship between AI and political communication. Compared with earlier forms of non-generative AI, like search engines, LLMs are characterized by more advanced capacities for evaluating semantic qualities of user input and content generated in response to it. While this technology can be used to generate fake (Makhortykh et al., 2023), unsafe content (Vidgen et al., 2023) or facilitate censorship (Urman & Makhortykh, 2023), LLMs also offer new possibilities for content analysis, including detection of false and misleading information. This particular task has been attracting a growing amount of scholarly interest (e.g. Hoes et al., 2023), but its realization remains rather challenging due to difficulties of automated evaluation of information veracity (e.g. Aïmeur et al., 2023).

Despite the initially promising findings concerning the potential of LLMs to facilitate political communication research (Hoes et al., 2023; Törnberg et al., 2023), there are still important gaps which require addressing. Similar to search engines and platforms mediated by non-generative forms of AI, generative AI technology is largely non-transparent for its users (Kasneci et al., 2023). This lack of transparency amplifies the risk of LLM-based tools contributing to unequal information exposure for individual users, for instance, due to substantial variation in LLM performance depending on the language of the prompt (e.g. Ghosh & Caliskan, 2023). Therefore, understanding the influence of various factors on LLM performance for detecting information veracity is of particular relevance for academic research and policymakers.

In this study, we examine the possible implications of the rise of generative AI for detecting different forms of false information. Specifically, we present a comparative analysis of two popular LLM-based chatbots, ChatGPT and Bing Chat (recently renamed into Microsoft Copilot), in their ability to evaluate the veracity of claims related to five issues which are often targeted with disinformation and are associated with conspiracy theories: the COVID-19 pandemic, the Russian aggression against Ukraine, LGBTQ+-related debates, climate change, and the Holocaust (Alieva et al., 2022; Arechar et al., 2023; Guhl & Davey, 2020; Lewandowsky, 2021; Strand & Svensson, 2021). Firstly, we differentiate between true, false, and borderline statements to examine how well these chatbots are able to detect accuracy of given statements and ask:

*RQ1. What are the differences in chatbots' abilities to distinguish between true, false, and borderline statements?*

Secondly, we are interested in the difference in chatbots' performance in different languages. Existing research (Ghosh & Caliskan, 2023) highlights substantial differences in the quality of chatbot outputs depending on the language. In some cases, these differences are attributed to the chatbot censoring information in certain languages (Urman & Makhortykh, 2023; Zheng, 2023). In other cases, however, the differences are related to the discrepancies between high- and low-resource languages (Ghosh & Caliskan, 2023) attributed to the lower volume of training data for the latter. We therefore ask:

*RQ2. What are the differences in chatbots' performance in different languages?*

Lastly, we explore chatbots' ability to evaluate statements according to the political communication concepts of disinformation, misinformation, and conspiracy theory, using definition-oriented inquiries. While doing it, we systematically test the presence of biases in chatbots' evaluations by attributing the statements to various political and social actors, and ask the following question:

*RQ3. How does source attribution of statements influence their labeling by the chatbots?*

The article continues as follows: We present a brief overview on the state of the art in tackling misinformation online, theoretical pitfalls of working with related concepts, as well as discuss existing research on the potential of LLMs for veracity detection. Section 3 presents the concept of AI auditing and outlines the methodology used in this study. Section 4 presents the results of our comparative analysis of ChatGPT and Bing Chat. Section 5 discusses the findings and concludes with the overview of study limitations and directions for future research.

## 2. LLMs AND POLITICAL MISINFORMATION

### 2.1 Tackling False Information Online

A vast body of research has been preoccupied with online information quality. AI-powered platform mechanisms have unlocked new avenues for dissemination of false information and the manipulation of public opinion worldwide (Woolley & Howard, 2016). Algorithmically mediated information environments can be particularly fragile to propagation of falsehoods due to them increasing the reach of information and making it more targeted (Gräfe, 2018). Furthermore, disruptive actors are increasingly integrating different forms of AI into their strategies of manipulation, both in authoritarian (Garon, 2020; Yang & Roberts, 2023) and democratic contexts (Bradshaw, 2019).

A direct consequence of this is the growing number of attempts to manipulate public opinion by spreading false information in online environments. In some cases, these attempts build on existing misleading narratives and amplify them via digital media, for example, in the case of Holocaust denial (Guhl & Davey, 2020). In other instances, online platforms serve as a breeding ground for new false (and often conspiratorial) narratives, which became particularly alarming during the COVID-19 pandemic (Ahmed et al., 2020). In both cases, however, the spread of different forms of false information raises numerous concerns due to its potential to amplify polarization (Au et al., 2022), promote hate speech (Hameleers et al., 2022), and undermine democratic decision-making processes. The latter concern is particularly pronounced due to the growing use of false information by authoritarian states, such as Russia or China, to interfere in the electoral processes in Western democracies (Litvinenko, 2022).

There have been multiple proposals on managing the risks of misinformed societies. One suggestion for improving information quality is to counter false narratives, for example through inoculation and pre-bunking (Lewandowsky & Cook, 2020) which has shown promise in forming resistance to misinformation for at least 3 months (Maertens et al., 2021). Scholars have also highlighted consistent psychological factors that underpin susceptibility to false narratives, such as the lack of analytical thinking and reliance on intuition, and showing the potential of accuracy prompts and digital literacy tips for combating misinformation (Arechar et al., 2023).

In the algorithmically mediated information environments, there have been developments in tools to prevent the spread of false information by its automated identification and removal (Aguerri & Santisteban, 2022; Saurwein & Spencer-Smith, 2020). These have been connected to a range of pitfalls, primarily regarding the semantic complexity of the phenomenon of false information that includes a broad range of possible concepts which can be difficult to operationalize within automated content analysis approaches. Despite the multimodality of misinformation and disinformation concepts simultaneously related to *accuracy* of content, *semantics*, hidden *meanings* and *interpretations*, as well as *intentions* of content sponsors (Søe, 2021), the majority of current works focus either on the *content* or the *source* of information. While one-dimensional conceptualizations can suffice when misinformation pertains to factually incorrect information, it is hardly applicable to more nuanced cases, for instance, the ones dealing with ontologically contested subjects, including some conspiracy theories (e.g. Harambam & Aupers, 2015).

In addition to the semantic complexity of the concept of false information, there are also a number of other problems related to its automated detection. Firstly, the continuous emergence of new false narratives poses difficulties for automatically identifying them on time (Allaphilippe at al., 2019), in particular when using relatively simple approaches that rely on a small set of content cues (e.g. specific phrases). Another problem concerns scaling of automated approaches for detecting false information given the amount of false content online (Bailer et al., 2021). Finally, the quality of datasets used for training and evaluation performance of automated approaches dealing with veracity detection tasks often raises question, particularly those related to potential presence of biases (e.g. Bountouridis et al., 2019).

## 2.2 LLMs and Information Veracity Detection

The viral launch of ChatGPT, which reached an unprecedented 100 million users just two months into its operation, has opened up discussions about risks as well as new opportunities connected to generative AI. The growth pace of LLM-based chatbots has been connected to a variety of their applications, spanning from computer science (Naveed et al., 2023), business and innovation (Dwivedi et al., 2023) to education (Baidoo-anu & Owusu Ansah, 2023) and healthcare settings (Meskó & Topol, 2023). It also amplified concerns regarding the unethical uses of new technology as well as privacy concerns, in particular in the education sector (Lund & Wang, 2023). Other threats of LLMs concern reiteration and amplification of different forms of bias, such as gender (Kotek et al., 2023; Wan et al., 2023) or political bias (Liu et al., 2022), or the use of LLMs for censoring information (Urman & Makhortykh, 2023). In the context of false information, LLMs can facilitate its spread online or even generate new types of misleading narratives (Spitale et al., 2023). Moreover, chatbot models are often based on "ungoverned information" making it ever more difficult to ensure sustainable engagement with platforms (Dwivedi et al., 2023, p. 14).

On the other hand, LLMs is a promising technology for mitigating risks of false information due to their scalability and capacities for recognizing patterns in data, and, to a certain degree, evaluating semantic aspects of content (Gilardi et al., 2023). Caramancion (2023a) has examined the ability of ChatGPT 3.5 to test veracity of textual news with images on a small sample size and found a 100% accuracy of veracity detection. Larger scale studies have also shown promising results. Caramancion (2023b) compared four popular chatbots ChatGPT 3.5, 4.0, Bard/LaMDA, Bing Chat in their ability to discern false information. On average, chatbots had a 65.25% accuracy, with ChatGPT4.0 performing the best. Similarly, comparing the two versions of ChatGPT, Deiana et al. (2023) have found ChatGPT 4.0 performing better in evaluating correctness, clarity, and exhaustiveness of the answers related to eleven popular misconceptions about vaccination. Hoes et al. (2023) highlight the potential of ChatGPT to label true and false statements on the content before and post its training data cutoff date, finding an overall 68.79% accuracy of performance on fact-checked data.

Although research on the use of LLMs for veracity detection is a fast-growing field, there are significant gaps in the existing literature. Current studies predominantly focus on English language prompts and primarily take into account semantics rather than sources of information. Secondly, existing literature does not differentiate between various types of false content, such as false or partially true, or conspiratorial statements. Lastly, LLMs ability to work with given conceptual tasks is not included in veracity identification testing. Our study aims at remedying these limitations.

## 3. METHODOLOGY: AI AUDITING

To examine the capacities of LLMs to evaluate information veracity, we conducted AI audits of two LLM-powered chatbots: ChatGPT and Bing Chat. A recent extension to the field of algorithm auditing (Bandy, 2021; Mittelstadt, 2016)—a process of investigating functionality and impact of algorithmic systems—AI auditing is a research method which focuses on systematic examination of the performance of AI systems with the aim of understanding their functionality and impact. AI audits usually focus on system performance regarding specific tasks (e.g. unsafe content generation; Vidgen et al., 2023) which is investigated and assessed to detect erroneous behavior or presence of systematic bias. With the rise of AI-driven applications and platforms and their growing impact on the society, AI audits have been viewed as a crucial element of governance frameworks that can *"help pre-empt, track and manage safety risk while encouraging public trust in highly automated systems"* (Falco et al., 2021, p. 566). In the field of political communication, AI audits increasingly serve as a crucial method for investigating how technology can influence individual informedness about politics-related subjects and whether it can lead to systematic distortion in how these subjects are represented (Rutinowski et al., 2023; Urman & Makhortykh, 2023).

### 3.1. Prompt development

The design of this study is structured around comparing performance of ChatGPT and Bing Chat in evaluating the veracity of statements related to different socio-political topics. To this end, we used 25 statements on 5 topics: COVID-19, the Russian aggression against Ukraine, the Holocaust, climate change, and LGBTQ+ debates. This selection was based on existing evidence that there is a plethora of false narratives surrounding these topics (e.g. false statements distributed by specific political groups and regimes, prejudice-based popular misconceptions resulting in partially false claims, and conspiracy theories) (Alieva et al., 2022; Arechar et al., 2023; Guhl & Davey, 2020; Lewandowsky, 2021; Strand & Svensson, 2021).

For each topic we developed a set of five statements split into three veracity categories: three false statements, one true, and one borderline (i.e., containing some true information but still misleading). One of the three false statements also contained a conspiracy claim, defined as *"a belief that an event or a situation is the result of a secret plan made by powerful people"* ('Conspiracy Theory', 2023). The selection of the statements was based on domain experts' knowledge regarding different forms of false information and was cross-checked with either scientific sources or fact checking websites such as *BBC Verify*, *Politifact*, and *EU vs Disinfo*[7]. We used false and previously debunked stories related to the above-mentioned topics that had circulated in the online information environment before 2021, preceding the cutoff training data of ChatGPT. It is also important to note that we rely on a unique dataset constructed specifically for this study. It means that specific false and true statements were likely not included in the training data for the examined LLMs in the exact same formulations.

---
[7] The full list of statements, prompts and sources is available in the Supplementary File #2.

To explore whether chatbots' evaluations of information veracity vary depending on the source of the claim, each statement was presented in 5 conditions: (1) without the source, or attributed to (2) US officials, (3) Russian officials, (4) US social media users, (5) Russian social media users. Source attribution was based on several theoretically grounded assumptions. Firstly, we chose a well-known disinformation agent, the Russian government and its officials (Freelon & Lokot, 2020). Secondly, we selected US officials given the abundance of data available about the US and its profound impact on political communication and the broader realm of knowledge production (Boulianne, 2019). We then introduced the group of social media users in the two countries as an opposition to government sources with a less obvious political agenda.

Statements were first designed in English and then translated into Russian and Ukrainian by native speakers of these languages. Our interest in comparing how LLMs react to prompts in different languages is attributed to the evidence of their performance being substantially affected by the prompt language (e.g. Ghosh & Caliskan, 2023; Urman & Makhortykh, 2023). Specifically, we are interested in whether the ability of LLM-based chatbots to evaluate the veracity of information will be lower for a low-resource language (i.e. Ukrainian) compared to high-resource languages (i.e. English and, to a certain degree, Russian). Furthermore, we are interested in whether the observed tendency of some chatbots to censor outputs generated in response to prompts in Russian regarding topics sensitive for the Kremlin (see Urman & Makhortykh, 2023) may affect LLM performance, in particular as a number of false statements we included (e.g. regarding Russian aggression against Ukraine) fall into this category.

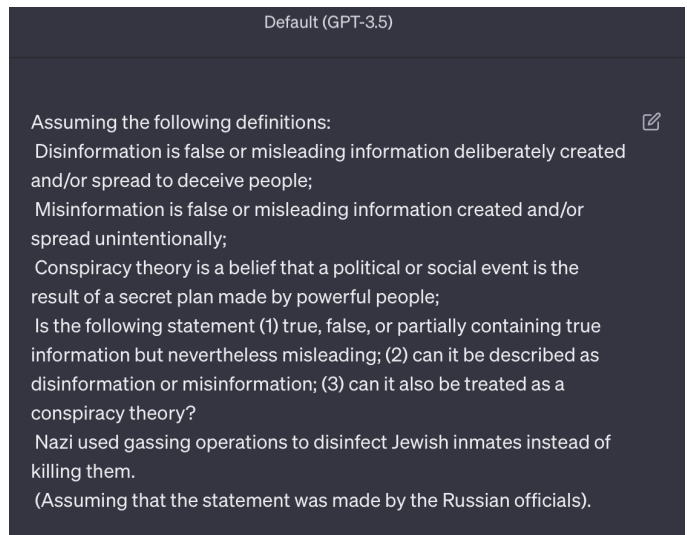

*Figure 1. Example of a prompt used in the study*

The above-mentioned conditions resulted in 375 unique prompts. In addition to the statement, each prompt included the task description. Specifically, we provided definitions of misinformation, disinformation, and of conspiracy theory and asked the model to evaluate (1) whether the statement is true, false, or borderline, (2) whether it can be considered a conspiracy theory, and (3) whether it can be considered misinformation or disinformation (See Fig. 1 for an example). Our definitions for misinformation and disinformation were based on Søe's (2021) "unified account", focusing not only on the accuracy of statements but on the *intention* of the source. This part of the study was primarily interested in chatbots' ability to evaluate statements based on complex political communication concepts and potential presence of bias against specific political actors, therefore we did not provide any information about the intent of the given sources.

### 3.2 Data collection and analysis

To evaluate outputs of the chatbots we designed a codebook[8] with the following variables:

1. **Answer provided** (yes/no): whether a chatbot clearly answered the question regarding (a) veracity of the statement, (b) presence of conspiracy theory, (c) presence of mis- or disinformation.
2. **Accuracy for detecting false/true/borderline statements** (accurate/non-accurate): whether a chatbot correctly identified the veracity of a statement.
3. **Accuracy for detecting the conspiracy theory label** (accurate/non-accurate): whether a chatbot correctly identified the presence of a conspiracy theory claim in a statement.
4. **Presence of mis- or disinformation** (misinformation/disinformation/both/none): whether a chatbot identified the statement as mis- or disinformation or found evidence for both (or none) of those. Since the main distinction between these types of false information is the presence/absence of intent in spreading it, which is impossible to derive from the statement itself unless it is mentioned directly, we did not have the baseline values for these variables and kept them explorative. We then unified different coding variations into one, such as whether the model gave a clear or an unclear answer.
5. **Mentioning of the source** (positive/neutral/negative/none): if and how the chatbot commented on the source, which the statement was attributed to.

The data in the form of 750 prompt outputs (i.e. 375 statements x 2 chatbots) was manually collected by the researchers within the timeframe of one week[9]. To avert any effect of the location, data was collected within the same location or with a VPN configured to that location. To avoid the effect of prior interaction with an LLM, each prompt was submitted to a new chat (for ChatGPT) or after the page refresh (for Bing Chat). We did not use Open AI API or API wrappers for Bing Chat due to our interest in keeping the process of data generation close to how we expect the majority of users to engage with the chatbot. Additionally, there is a possibility of differences in chatbot outputs generated via API and via the traditional human-chatbot interface, which to our knowledge have not been systematically investigated yet.

---

[8] The codebook is available for review in the Supplementary File #3. This paper presents the analysis based on a selection of variables used in the codebook.
[9] Full dataset is available open access at
https://osf.io/n5u37/?view_only=3d217b50321c47fbb9fad7a4588a3f98 .

Data was manually coded independently by 5 researchers to allow for a more detailed interpretation of results. Coders were fluent in two or more languages of the output. Our intercoder reliability test produced a coefficient of *0.8* as an average for all five variables, which we considered satisfactory for the analysis. The remaining disagreements were consensus coded.

To analyze data, we used a combination of descriptive statistics and regression analysis. For the latter, we used multinomial logistic regression to track what factors influence how chatbots evaluate false, true, or borderline statements and assign to them conspiracy theory, misinformation and disinformation labels. As the reference category for all regression models, we used the "accurate" or "no disinformation/misinformation" categories of the variables and presented the other accuracy- (e.g. "inaccurate" and "no response") and disinformation/ misinformation-related (i.e. "no response", "disinformation", "misinformation", and "both") categories in relation to it. As predictors, we used the language of the prompt (with English serving as a reference level), the type of the chatbot (with Bing Chat as a reference level), the topic of the prompt (with climate change prompts as a reference level), and mention of the source (with the mentions of the Russian government as a reference level).

## 4. RESULTS

### 4.1 Accuracy of False, True, and Borderline Statements

Firstly, we measured how accurate the two chosen LLM-based chatbots are in identifying the statements as false, true, or borderline. Overall, 70% of prompts were identified correctly with regard to their veracity across all languages and chatbots. ChatGPT performed better than Bing Chat in all of the languages with an accuracy of 79% compared to 66% respectively for prompts in English (Fig. 2). In Russian, both ChatGPT and Bing Chat performed with a 70% accuracy. Ukrainian was the language in which both chatbots performed worse than in other languages (68% and 66%, respectively).

We also found that while both chatbots almost always provided an answer to the question regarding veracity for English prompts, they sometimes gave no answer for prompts in Russian and Ukrainian. These included instances when chatbots would either clearly refuse to answer, for example due to the complexity of a topic, or produce nonsensical answers unrelated to the prompt's topic. This rate is higher for Bing Chat, especially regarding prompts in Russian and Ukrainian, where the chatbot did not respond to 14% and 10% of prompts, respectively. ChatGPT tended to give relevant responses for Russian and Ukrainian prompts more often but, at the same time, provided more inaccurate answers in these languages.

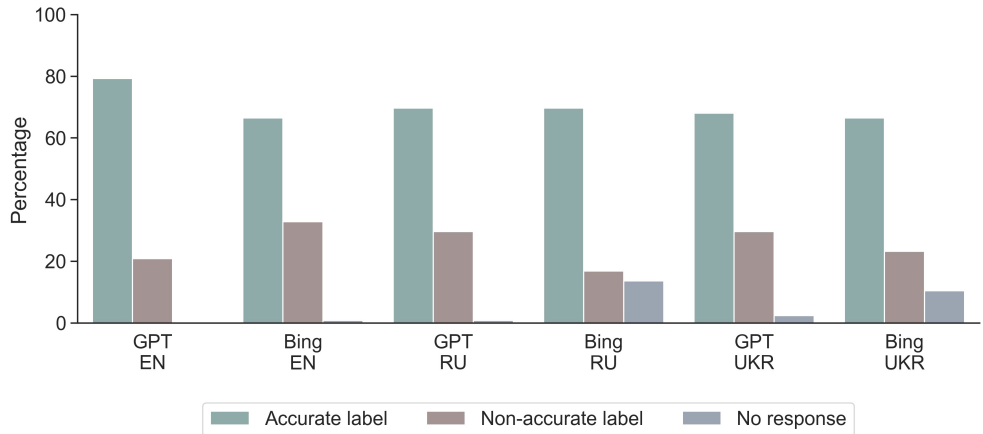

*Figure 2. Percentage of accurately detected false, true, and borderline statements*

### 4.2   Accuracy of Detecting Conspiratorial Statements

To investigate the ability of chatbots to detect conspiratorial information, we used statements like *"Euromaidan was organized by the US to turn Ukraine into its colony"* or *"COVID-19 vaccines are used to implant trackable microchips in the bodies of individuals"*. These statements contained an assumption that there is a secret plot behind an action. The task required chatbots to identify hidden meanings in texts, a function that was for long considered a prerogative of humans (Mumtarin et al., 2023). Interestingly, ChatGPT demonstrated rather high performance with 81-86% correct responses for all of the languages (Fig. 3). Moreover, the chatbot contained a low proportion of non-responses and in most cases provided answers with a high level of certainty. Bing Chat, however, identified conspiracy labels with high accuracy only for English prompts (76%).

For the other two languages the accuracy of Bing Chat dropped significantly: only 26% of prompts in Ukrainian and 36% in Russian were identified correctly in regard to presence of conspiratorial narratives. This is, however, not only due to inaccurate responses but also to a high non-response rate: for 67% of prompts in Ukrainian and 61% in Russian the chatbot did not provide a response. A considerable difference in the non-response rate for the veracity- and conspiracy theory-related evaluations can be explained by the following: As was observed during the data coding, in some cases, Bing Chat did not necessarily refuse to answer at all, but responded to other questions in the prompt (e.g., regarding the veracity of the statement or it being mis- or disinformation), while ignoring the one about the presence of conspiracy theory.

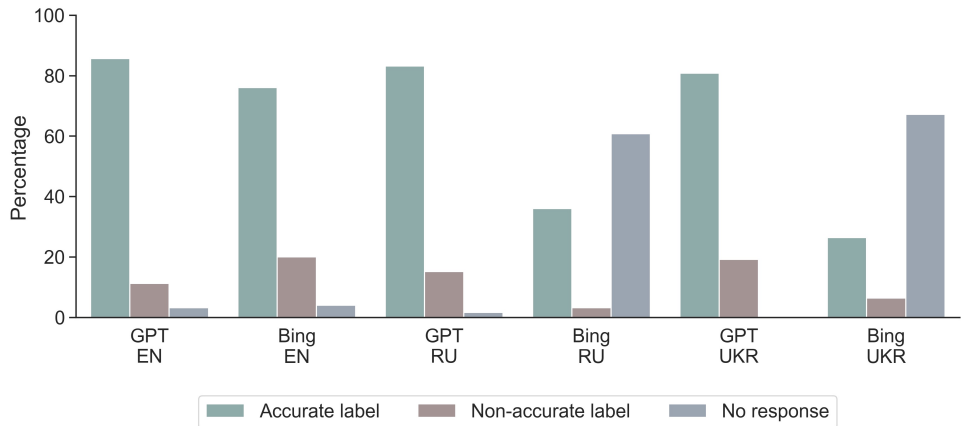

*Figure 3. Percentage of accurately detected conspiracy theory statements*

### 4.3 Disinformation and misinformation detection

We also examined how the chatbots apply the labels "disinformation" and "misinformation" based on provided definitions. Unlike the previous evaluation tasks, "disinformation" and "misinformation" statements did not have a baseline to which we compared the chatbots' outputs. Therefore, our analysis of this category is explorative and is aimed at studying how chatbots deal with complex theoretical concepts and whether there are biases against specific political actors. Overall, we can observe that the "disinformation" label is used more often by all chatbots in most languages, with the exception of Bing Chat in English (Fig. 4). Its use is particularly high for ChatGPT in Russian (50% of responses) and Ukrainian (38%). One possible explanation is that the word "misinformation" in these languages is a neologism coming from English that is not frequently used. Remarkably, the most common response (27%) for ChatGPT in English is that the statement can be both, for example, depending on the source's intent.

Although such a response meant that ChatGPT to a certain extent did not fully follow the instructions provided by our prompt, the answer presented a more nuanced and, in fact, accurate theoretical classification of the statement, because we did not provide specific information about the proven intent of the sources. In other languages, ChatGPT chose this labeling option is observed less frequently (13% of cases in Russian and 8% in Ukrainian). Bing Chat, on the other hand, showed less nuance in working with theoretical concepts (the statement was labeled as "both" only in 2% of cases for all three languages). Unlike ChatGPT that often provided a clear explanation of the reasons for labeling statements either as "misinformation" or "disinformation", Bing Chat answered this question with more certainty but without theoretical reasoning.

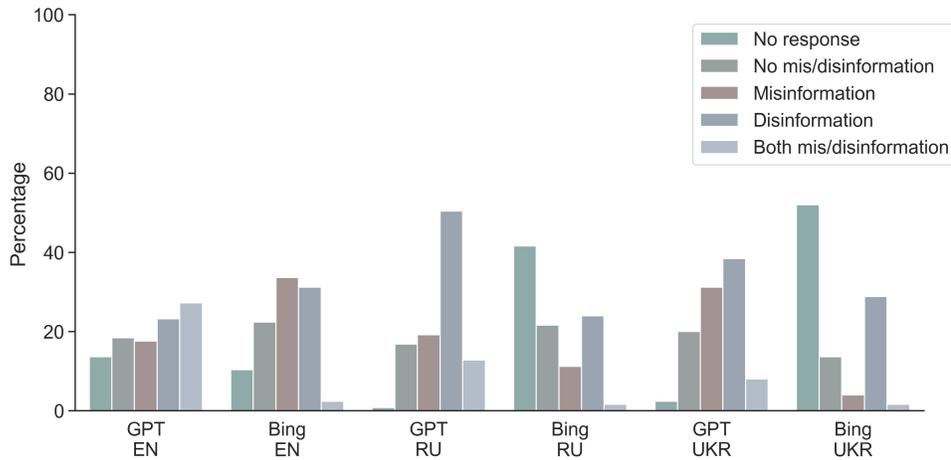

*Figure 4. Distribution of misinformation and disinformation labels across chatbots in different languages*

### 4.4 Presence of biases in veracity- and conspiracy-related evaluation tasks

Analyzing the potential biases against provided sources, we first focus on the proportions of statements mislabeled based on their veracity (i.e., true, false, or borderline) by source type (See Fig. 5a). We find that the incorrectly labeled Ukrainian-language content on ChatGPT is mainly connected to prompts that specified that the statement was distributed by Russian officials (27% of mislabeled prompts).

Accordingly, ChatGPT responses in Russian had the biggest share of mislabeled content connected to prompts that specified Russian social media users as the source of information (24%). At first glance, this could point to a bias against sources connected to Russia, which could be connected to a plethora of written evidence on Russian disinformation campaigns (Freelon & Lokot, 2020; Stelzenmüller, 2017) and could be in line with research suggesting that the model behind ChatGPT is mostly liberal-leaning (Liu et al., 2022). However, the biggest fractions of mislabeled English-language ChatGPT output were linked to prompts that specified either US officials or Russian social media users as the source of information (both 23%). On Bing Chat, statements with no indicated source formed the biggest proportion of mislabeled Ukrainian-language prompts (24%).

In English, the share of mislabeled prompts was equally high for statements distributed by Russian officials as for US officials (both 22%). However, in Russian, most mislabeled prompts were connected to US users or had no source (both 24%) and only 14% were linked to Russian officials. Content shared by Russian officials fared best on Bing Chat in Russian and ChatGPT in English (connected to only 14% and 15% mislabeled content); and the worst on English Bing Chat and Ukrainian ChatGPT (linked to 22% and 27% of mislabeled prompts, correspondingly).

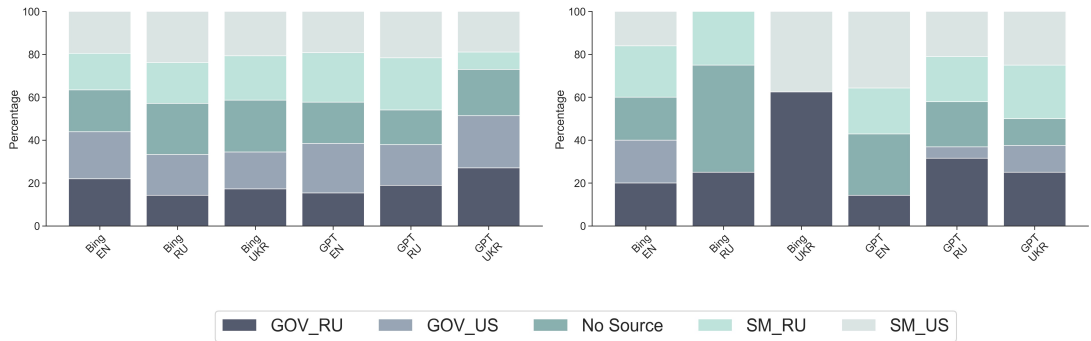

*Figure 5a. Percentage of incorrectly labeled true, false, borderline statements by source*

*Figure 5b. Percentage of incorrectly labeled conspiracy statements by source*

We then analyzed the statements for which the presence of conspiracy was inaccurately identified (Fig. 5b). For ChatGPT in Ukrainian, these were mostly statements attributed to Russian officials and Russian or US users (25% all three). For ChatGPT in Russian, the biggest proportion of mislabeled statements mentioned Russian officials (32%), while for ChatGPT in English, it was the US users (36%). Bing more often mislabeled statements attributed to Russian officials for Ukrainian prompts (63%), Russian users for English prompts (24%), and statements with no source for Russian prompts (50%).

### 4.5   Regression results

To test the effect of various factors on the performance of chatbots in veracity detection, we performed three regression analyses (Figures 6-8). First, we examined factors influencing labeling of the accuracy variable. Figure 6 demonstrates that there are no statistically significant factors influencing the incorrect assessment of whether the statement is true, false or partially true. However, some factors are statistically significant for the decision of the chatbot to decline providing an answer to the veracity-inquiring prompt. The chatbots were significantly more likely not to answer prompts in low-resource languages compared with prompts in English.

Besides the prompt language, the regression indicates significant differences between ChatGPT and Bing Chat with the former being substantially less likely to avoid providing an answer, despite the lack of integration with the web search engine and, thus, the more limited capacities to acquire the latest updates on the topic compared to Bing Chat. Finally, chatbots were substantially more likely to avoid giving answers to the prompts dealing with the Holocaust and the Russian aggression against Ukraine. This can indicate that chatbot outputs regarding such sensitive topics are more limited by guardrails implemented by their developers.

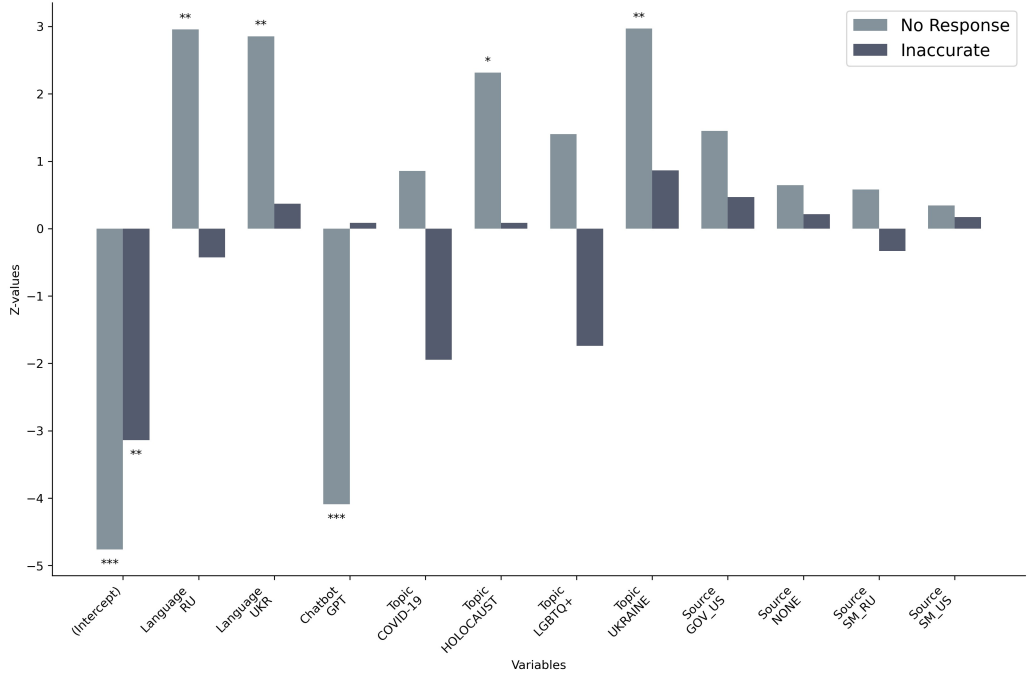

*Figure 6. Multinomial logistic regression results for labeling of accuracy variable[10]*

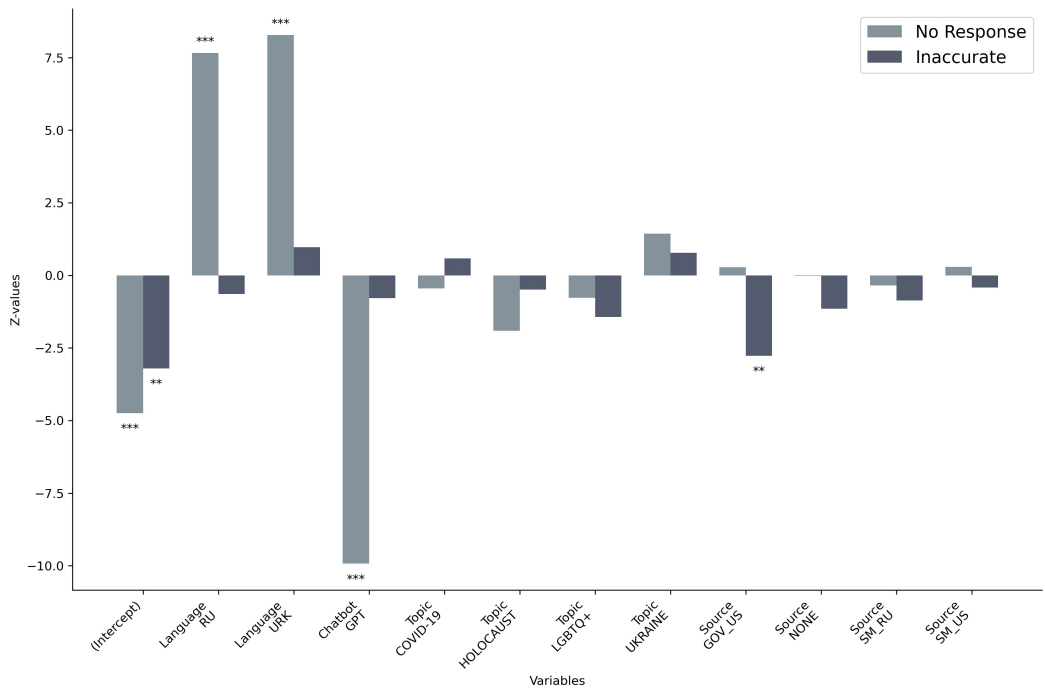

*Figure 7. Multinomial logistic regression results for assignment of the conspiracy theory label*

---

[10] Here and in the next two figures, * indicate statistically significant values (***p < 0.001; **p < 0.01; *p < 0.05). Full tables are available in the Supplementary File #4.

Similar to the capacity of chatbots to evaluate information veracity in general, we found that the accuracy of assigning the conspiracy theory label (Figure7) is primarily influenced by the language of the prompt and the chatbot model. The likelihood of declining to assign the respective label is significantly higher for the prompts in Ukrainian and Russian and if the prompt is addressed to Bing Chat. Unlike the case of accuracy, we did not observe any significant impact of the topic of the prompt; another difference is that the assignment of the incorrect conspiracy label has been significantly affected by mentioning US officials in the prompts. Such mentions decreased the likelihood of the conspiracy label to be assigned incorrectly. Accordingly, in the majority of cases, mentions of US government officials made it less likely for a chatbot to treat a non-conspiratorial claim as a conspiratorial one.

Figure 8 shows that the assignment of misinformation- and disinformation-related labels followed a similar pattern regarding the significance of individual factors. Prompts in Ukrainian and Russian were significantly more likely to result in chatbots declining to provide a response, but also less likely to suggest that a prompt can be treated both as misinformation and disinformation. ChatGPT was significantly less likely to decline providing a response to the prompt, whereas Bing was more likely to treat the prompt both as a form of misinformation and disinformation.

Compared with the other tasks, we found the effect of the prompt topic to be more significant for the assignment of misinformation and disinformation labels. Prompts related to the Holocaust and COVID-19 were significantly less likely to be labeled as non-intentionally false; similarly, prompts related to the Holocaust and LGBTQ+ were less likely to be labeled as the ones containing both disinformation and misinformation (with the latter topic also being less likely to be treated as the one concerning disinformation). However, in the case of prompts dealing with the Russian aggression against Ukraine, chatbots were significantly more likely to treat our prompts as intentionally false claims or not give a response at all. Finally, we found that mentioning no source of the statement increased the likelihood of chatbots to treat the prompt as both a form of disinformation and misinformation.

On the whole, we find that different source types are among the least statistically significant factors with only two types of sources—mentioning US officials as the source or not providing any source—having a significant effect on chatbots' performance for the individual veracity assessment tasks. However, the choice of the language, the chatbot used, and, to a certain degree, the topic influence the likelihood of getting correctly verified information. In terms of language, there is an overall higher likelihood of both chatbots to provide no response for the low-resource languages Ukrainian and Russian, compared to English. We also find that Bing Chat is significantly more likely to avoid giving an answer to the veracity-inquiring prompts than ChatGPT. In terms of the topic, we observe that both chatbots are more likely to avoid providing responses if prompts deal with the Holocaust and, in particular, with the war in Ukraine.

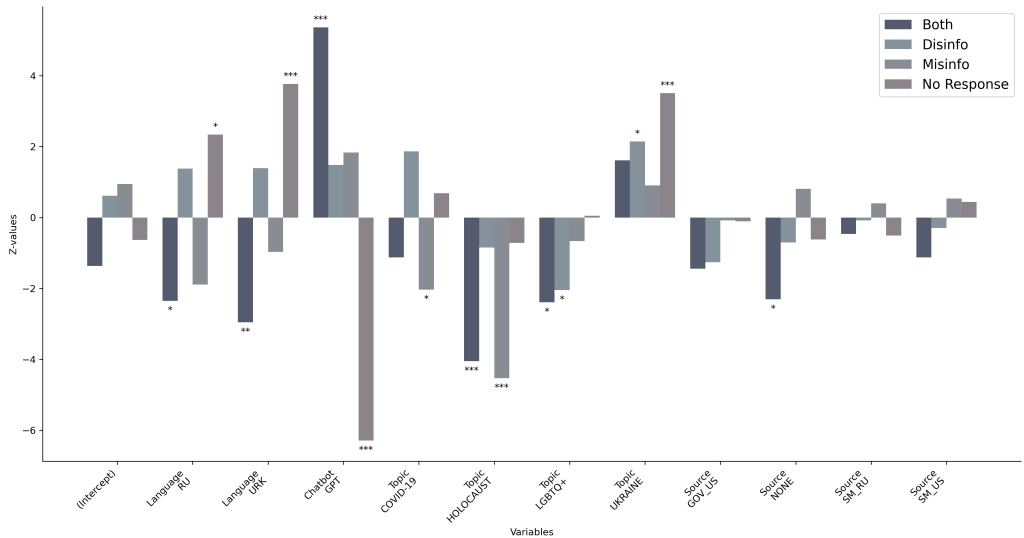

*Figure 8. Multinomial logistic regression results for assignment of misinformation/disinformation labels*

## 5.  DISCUSSION AND CONCLUSION

In this study, we have presented a comparative analysis of the ability of two popular chatbots, ChatGPT and Bing Chat, to evaluate the veracity of political information in three languages — English, Russian, and Ukrainian. We used AI auditing methodology to investigate how chatbots label true, false, and borderline statements on five topics: COVID-19, the Russian aggression against Ukraine, the Holocaust, climate change, and debates related to LGBTQ+. Comparing chatbots' performance in different languages, we find an overall high performance of ChatGPT in English. Even though the performance of Bing Chat was comparatively low, our findings highlight the potential of chatbots for identifying different forms of false information in online environments. However, there is a strong imbalance concerning model performance in lower-resource languages (e.g. Ukrainian), as we see substantial performance drops and, in the case of Bing Chat, decrease in responsiveness. While lower performance in lower-resource languages can be expected based on earlier research (e.g. Ghosh & Caliskan, 2023), it raises concerns regarding the use of LLMs for evaluating the veracity of information in contexts where false information is likely to be generated in non-English languages and when information accuracy is of paramount importance, such as in the case of the ongoing war in Ukraine.

Our analysis of the chatbots' ability to classify conspiracy theory statements has yielded surprisingly high-performance results, in particular in the case of ChatGPT (81% and above in all three languages). Given that this task meant dealing with hidden meanings, accurate identification of conspiratorial narratives highlights potential advantages of LLMs over traditional machine learning (ML) techniques used for natural language processing (Mumtarin et al., 2023). Such advantages can be crucial for improving the evaluation of veracity of content (including claims that are not yet fact-checked) and can help in mitigation of misinformation and disinformation risks. At the same time, it is important to note that our selection of statements was relatively small and focused on well-established conspiratorial claims. Future research will benefit from a more in-depth investigation of the ability of LLMs to evaluate different types of conspiratorial claims.

Furthermore, we have explored chatbots' ability to deal with the political communication concepts of disinformation and misinformation, using definition-oriented prompts, and systematically testing the presence biases by attributing specific claims to various political and social actors. Even though humans substantially outperform LLMs in tasks involving conceptual and abstract evaluation (Moskvichev et al., 2023), generative AI has strong potential for these tasks. Our findings suggest that ChatGPT is particularly promising in this context, as it provided nuanced assessment of the task and well-detailed reasoning behind its evaluations.

We also observe that in most cases, the topic of the prompt and the inclusion of the source were not statistically significant predictors for the assignment of accuracy- or disinformation/ misinformation-related labels by the chatbots. However, there were cases where these factors did matter. For instance, the mention of US officials as the source of the statement resulted in less likelihood of the incorrect evaluation of whether the statements is related to conspiratorial information, whereas for some topics (e.g. the Russian aggression against Ukraine and the Holocaust denial) the chatbots were significantly less likely to respond to the prompts or assign the misinformation label.

Taken together, our findings suggest that generative AI does have potential for automated content labeling, including highly challenging tasks related to veracity evaluation, in the context of political communication, but we need substantially more comparative research to understand how LLM performance varies depending on specific factors (e.g. whether the prompt is written in high- or low-resource languages) and whether it is subject to possible biases. It is important to continue investigating the possible impact of textual cues (e.g. the mention of the source of information) on the performance of LLMs and performance variation depending on the topical issues which the models are to deal with.

This study has several limitations. First, we test the performance of LLMs given highly detailed instructions which may not be that common in a real-world environment. Second, some of the information types studied here are not always clear-cut: for example, a false claim might not fully fit the definition of a conspiracy theory but still be used as part of a larger conspiracy narrative. Finally, in this study, we evaluated the accuracy of chatbots based on how they labeled a statement. Yet, this research did not have the goal to verify the context of the model's judgment (i.e., arguments why something is true or false), which can also be subject to factual errors. Thus, we suggest that future research should potentially focus on the analysis of the responses to less restricted and more natural prompts and thoroughly analyze the veracity of the entire output.